\documentclass[10pt,twocolumn,letterpaper]{article}

\usepackage{cvpr}
\usepackage{times}
\usepackage{epsfig}
\usepackage{graphicx}
\usepackage{amsmath,amsfonts,amssymb}
\usepackage{amsxtra}

\usepackage{multirow}
\usepackage{threeparttable}
\usepackage{graphicx}
\usepackage{bbm}
\usepackage{color}
\usepackage{booktabs} 
\usepackage{times}
\usepackage{subfig}
\usepackage{mathrsfs}

\newcommand{\matr}[1]{\mathbf{#1}}

\usepackage[pagebackref=true,breaklinks=true,letterpaper=true,colorlinks,bookmarks=false]{hyperref}

 \cvprfinalcopy 

\ifcvprfinal\fi
\begin{document}

\title{C3AE: Exploring the Limits of Compact Model for Age Estimation}

\author{Chao Zhang$^{1,2}$, Shuaicheng Liu$^{1,2}$, Xun Xu$^3$, Ce Zhu$^{1,}$\thanks{Corresponding author}\\
University of Electronic Science and Technology of China$^1$
Megvii Technology$^2$\\National University of Singapore$^3$\\
{\tt\small galoiszhang@gmail.com, liushuaicheng@megvii.com, eczhu@uestc.edu.cn, elexuxu@nus.edu.sg}
}

\maketitle

\begin{abstract}
Age estimation is a classic learning problem in computer vision. Many larger and deeper CNNs have been proposed with promising performance, such as AlexNet, VggNet, GoogLeNet and ResNet. However, these models are not practical for the embedded/mobile devices. Recently, MobileNets and ShuffleNets have been proposed to reduce the number of parameters, yielding lightweight models. However, their representation has been weakened because of the adoption of depth-wise separable convolution. In this work, we investigate the limits of compact model for small-scale image and propose an extremely \textbf{C}ompact yet efficient \textbf{C}ascade \textbf{C}ontext-based \textbf{A}ge \textbf{E}stimation model(\textbf{C3AE}). This model possesses only 1/9 and 1/2000 parameters compared with MobileNets/ShuffleNets and VggNet, while achieves competitive performance. In particular, we re-define age estimation problem by two-points representation, which is implemented by a cascade model. Moreover, to fully utilize the facial context information, multi-branch CNN network is proposed to aggregate multi-scale context. Experiments are carried out on three age estimation datasets. The state-of-the-art performance on compact model has been achieved with a relatively large margin. 
\end{abstract}

\section{Introduction}
Convolutional neural networks (CNNs) are being developed deeper and larger for more precise accuracy in recent years. This trend has brought in unprecedented computation cost to either training or deploying. In particular, deploying existing classic large models, e.g., AlexNet~\cite{krizhevsky2012imagenet}, VggNet~\cite{simonyan2014vgg} and ResNet~\cite{Kaiming2015Deep}, on mobile phones, cars and robots is next to impossible due to the model size and computational cost. 

To deal with above problem, recently MobileNets~\cite{howard2017mobilenets, sandler2018mobilenetv2} and ShuffleNets~\cite{zhang1707shufflenet, ma2018shufflenet} have been proposed to greatly reduce the parameters by exploiting the depth-wise separable convolution. In these models, the traditional convolution is replaced by two step convolutions, namely the filtering layer and combining layer. For example, in MobileNets, the filtering layer first convolves each  corresponding channel separately, thus breaking the interactions among various output channels, which can reduce the number of parameters dramatically. A $1\times1$ convolution then stitches different channels to combine the information acquired from different input channels. For large-scale images, such operation is reasonable because images need to be represented by large number of channels, e.g., 512 and 384 in VggNet~\cite{simonyan2014vgg} and ResNet~\cite{Kaiming2015Deep}. Whereas, for small-scale images, e.g., images with low resolution and small dimension, such predicate remains questionable. 

\begin{figure}[t]
	\centering
	\includegraphics[width=0.47\textwidth]{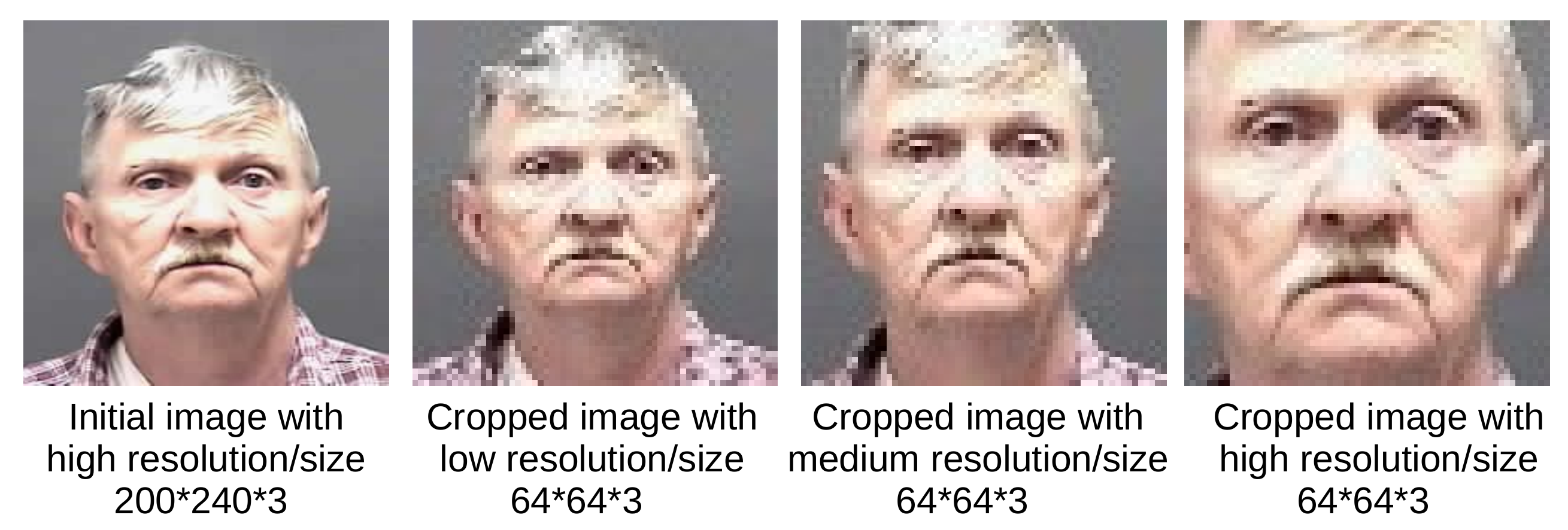}\\
	\caption{Human can recognize the age of person in one of the four images, regardless of different resolutions or scales. Is it necessary to use the first image that is with large size? In this work, we use small-scale image ($64 \times 64 \times 3$) for age estimation, which can achieve very  competitive performance.}
	\label{Fig: motivation}
\end{figure}

In contrast to large-scale images, small-scale images can be often represented by fewer number of channels in the network, and so does the number of parameters and memory. Therefore, standard convolution layer with small size kernel does not require much more parameters and memory compared with depth-wise separable convolution \cite{howard2017mobilenets,zhang1707shufflenet}. 
From the perspective of image representation, the output channels of depth-wise convolution are many times larger than that of standard convolution. To compensate the representation ability, the depth-wise convolution has to pay for the cost of increased parameters. Therefore, we believe the conventional convolution layer with small kernel size is more suitable for processing small-scale images than depth-wise counterpart.

Images must often be stored and processed with low resolution and scale, aka small-scale images, on low-cost mobile devices. One of the eminent problems which falls into the category is age estimation. For example, human can easily recognize the age of the man in Fig.~\ref{Fig: motivation} in either full or low resolution and partial or full view of the face. We, therefore, conjecture such ability is applicable to contemporary CNNs and design a compact with standard convolution layers with small-scale face images as input for age estimation.

Recent advances in age estimation are usually summarized into two mainstream directions: jointly category classification and value regression, and distribution matching. For the former, the psychological evidence \cite{Hutchins} reveals that humans are inclined to give categorical ratings on image rather than continuous scores, i.e., preferring to different levels. Some works \cite{liu2017ordinal, feng2017human} utilize the category information and ordinal information to implement classification and regression simultaneously. For the latter one, distribution matching can achieve promising results under the assumption that distribution label of each image is provided. Nevertheless, acquiring distributional labels for thousands of face images itself is a non-trivial task. In this work, we propose to exploit the information on classification, regression and label distribution simultaneously. This is achieved by representing discrete age as a distribution over two discrete age levels and the training objective is to minimize the match between distributions.
In deep regression model, a fully connected layer with semantic distribution is inserted in between the feature layer and age value prediction layer.

To summarize, we design a compact model that takes small-scale image as input. 
Specifically, we utilize standard convolution instead of depth-wise convolution, with suitable kernel and number of channels. To the best of our knowledge, this is the smallest model that has been obtained so far on the facial recognition, i.e., 0.19MB for plain model and 0.25MB for full model. We then represent the discrete age value as a distribution and design a cascade model. Moreover, we introduce a context based regression model which takes as input multiple scales of facial image. With the \textbf{C}ompact basic model, \textbf{C}asaced training and multi-scale \textbf{C}ontext, we aim to tackle small-scale image \textbf{A}ge \textbf{E}stimation. Thus we name the network \textbf{C3AE}.

Our main contributions are as follows. 
First, we study the relationship between the channel number and the representation on depth-wise convolution, especially on the small scale image. 
Our discussion and results advocate a rethinking of MobileNets and ShuffleNets for the small-scale/medium-scale images. Second, we present a novel age representation that exploits the information on classification, regression and label distribution simultaneously and design a cascade model. Finally, we propose a context based age inference method collecting different granularity of input images. The proposed model, named C3AE, achieves the state-of-the-art performance compared with alternative compact models and even outperforms many bulky models. With the extremely compact model (0.19MB and 0.25 MB for plain and full model, respectively), C3AE is suitable to be deployed on low-end mobiles and embedded platforms.

\section{Related Work}
\noindent\textbf{Age Estimation}
The age progression displayed on faces is uncontrollable and personalized~\cite{fu2010age}, and the traditional methods 
often have the problem of generalization. With the success of deep learning, many recent works applied deep CNN to achieve the state-of-the-art performance on various applications such as image classification~\cite{krizhevsky2012imagenet, simonyan2014vgg, szegedy2015going, szegedy2016rethinking, Kaiming2015Deep, szegedy2017inception, huang2017densely}, semantic segmentation~\cite{long2015fully, chen2018deeplab}, object detection~\cite{girshick2014rcnn, ren2015faster_rcnn, redmon2017yolo9000}. As for age estimation, 
CNNs are also being used for its strong generalization. Yi~\etal \cite{yi2015age} firstly utilized CNN models to extract features from several facial regions, and used a square loss for age estimation.
AgeNet~\cite{levi_ageAndgender} used one-dimensional real-value as an age group for age classification. 
Rothe~\etal~\cite{Rothe2016Deep} proposed to use expected value on the softmax probabilities and discrete age values for age estimation. 
It is a weighted softmax classifier only in the testing phase. 
Niu~\etal~\cite{niu2016ordinal} formulated age estimation as an ordinal regression by employing multiple output CNNs. Following ~\cite{niu2016ordinal}, Chen~\emph{et al.}~\cite{chen2017using} utilized ranking-CNN for age estimation, in which there were a series of basic binary CNNs, aggregating to the final estimation.
Han~\etal~\cite{Han2018Heterogeneous} used multiple attributes for multi-task learning.
Gao~\etal~\cite{gao2017deep} used KL divergence to measure the similarity between the estimated and groundtruth distributions for age. Pan~\etal~\cite{pan2018mean} designed a new mean-variance loss  for distribution learning.

However, in real applications, the distribution is usually not available for a face image. In this work, we consider two objectives simultaneously. The first one minimizes the Kullback-Leibler loss between distributions, and the second one optimizes the squared loss between discrete ages. 

\noindent\textbf{Compact Model}
As the increasing requirement of mobile/embedded devices running deep learning, various efficient models such as GoogLeNet~\cite{szegedy2015going}, SqueezeNet \cite{iandola2016squeezenet}, ResNet~\cite{Kaiming2015Deep} and SENet~\cite{hu2017squeeze}, are designed to cater this wave.
Recently, depth-wise convolution was adopted by MobileNets~\cite{howard2017mobilenets, sandler2018mobilenetv2} and ShuffleNets~\cite{zhang1707shufflenet, ma2018shufflenet} to reduce computation costs and model sizes. They were built primarily from depth-wise separable
convolutions initially introduced in \cite{sifre2014rigid} and subsequently used in Inception models \cite{szegedy2016rethinking, szegedy2017inception} to reduce the computation in the first few layers. In particular, the separation of filtering - applying convolution at each channel separately and combination - recombine the output of individual channels achieved fewer computations.
MobileNet-V1~\cite{howard2017mobilenets} based on the depth-wise separable convolution explored some important design guidelines for an efficient model. 
ShuffleNet-V1~\cite{zhang1707shufflenet} utilized novel point-wise group convolution and channel shuffle to reduce computation cost while maintaining accuracy. 
 MobileNet-V2~\cite{sandler2018mobilenetv2} proposed a novel inverted residual with linear bottleneck.
ShuffleNet-V2~\cite{ma2018shufflenet} mainly analyzed the runtime performance of the model and give four guidelines for efficient network design. 

For age estimation, we argue that for small-scale images, the channel size is often small and the depth-wise separation does not benefit. Instead, a standard convolution is adequate for the trade-off between accuracy and compactness.

\section{The Proposed Model}
In this section, we firstly present the compact model and its architecture as well as some important discussions on practical guidelines. Then we describe a novel two-points representation of age, and utilize the cascade style to insert it in deep regression model. Next a context based module is embedded into a single regression model by exploiting facial information at three granularity levels. Finally some discussions are given for rethinking.

\subsection{Compact Model for Small-scale Image: Revisiting Standard Convolution}
Our plain model is composed of five standard convolution and two fully connected layers as shown in Tab.~\ref{Tab: architecture}\footnote{(-) in the whole manuscript indicates value not available, or also useless for comparison.}. For standard convolution layer followed by batch normalization, Relu and average pooling, its kernel, number of channels and parameters are 3, 32 and 9248, respectively. 
As a basic module, we will show why we use standard convolution block instead of the separable convolution block that used in MobileNets and ShuffleNets.
We shall demonstrate later in the experiment, our basic model produces competitive performance compared with fashionable models though its simplicity. 

\begin{table}[!htb]
	\footnotesize
	\begin{center}
		\caption{Overall architecture of the compact plain model}
		\tabcolsep=2pt
		\begin{tabular}{c | c  c  c  c  c  c}
			\hline
			\hline
			Layer  & Kernel & Stride & Output size & Parameters & MACC \\
			\hline
			Image  & -      & 1      & 64*64*3     & -         &  -     \\
			\hline
			Conv1  & 3*3*32 & 1      & 62*62*32    & 896       &  3321216    \\
			\hline
            BRA    & -      & 1      & 31*31*32    & 128        &  -      \\
			\hline
			Conv2  & 3*3*32 & 1      & 29*29*32    & 9248       &  7750656     \\
			\hline
            BRA    & -      & 1      & 14*14*32    & 128       &  -     \\
            \hline
			Conv3  & 3*3*32 & 1      & 12*12*32    & 9248       &  1327104    \\
			\hline
            BRA    & -      & 1      & 6*6*32      & 128       &  -     \\
			\hline
			Conv4  & 3*3*32 & 1      & 4*4*32      & 9248      &  147456    \\
			\hline
           BN+ReLu & -      & 1      & 4*4*32      & 128        &  -     \\
			\hline
		   Conv5   & 1*1*32 & 1      & 4*4*32      & 1056     &  16384     \\
		   \hline
		   Feat    & 1*1*12 & 1      & 12          & 6156       &  -    \\
		   \hline
		   Pred    & 1*1*1  & 1      & 1           & 13        &  -    \\
		   \hline
		   \hline
		   Total   & -      & -      & -           & 36377     & -   \\
		   \hline
		\end{tabular}
		\begin{tablenotes}
			 (BRA) indicates batch normalization(BN), Relu and average pooling.\\
			  (MACC) Here we only count MACC of the conv layer.
		\end{tablenotes}  
		\label{Tab: architecture}
	\end{center}
\end{table}

In MobileNets, the status regarding the saving of parameters and computation were analyzed, especially comparing between standard convolution and depth-wise separable convolution. That analysis is suitable for large-scale image while for the small-scale/medium image it may not work well. 

Given an input and output as $D_F \times D_F \times M$ feature map \textbf{F} and $D_F \times D_F \times N$ feature map $\textbf{G}$, $D_F$ denotes the size of feature map, $M$ and $N$ are the number of input channels and output channels for a convolution layer, respectively. 
The number of computation cost is given by $D^2_K \cdot M \cdot D^2_F + M \cdot N \cdot D^2_F$ \cite{howard2017mobilenets}. In comparison, the standard convolution layer is parameterized by convolution kernel \textbf{K} of size $D^2_K  \times \hat{M} \times \hat{N}$. The reduction between  standard convolution and depth-wise separable convolution in computation cost \cite{howard2017mobilenets} is:
\begin{equation}\label{eq:saving_experss}
\resizebox{.8\hsize}{!}
{$
\begin{split}
&\frac{D_K^2 \cdot M \cdot D_F^2 + M \cdot N \cdot D_F^2}{D_K^2 \cdot \hat{M} \cdot \hat{N} \cdot D_F^2}=\frac{M}{\hat{M}\hat{N}} + \frac{MN}{\hat{M}\hat{N}D^2_K}
\end{split}
$}
\end{equation}
Only with the assumption that both the depth-wise convolution and standard convolution need the same channel size, i.e. $M=\hat{M}$ and $N=\hat{N}$, Eq.~\ref{eq:saving_experss} can be reduced to $\frac{1}{N} + \frac{1}{D^2_K} < 1$. However, the depth-wise convolution often requires much more channel numbers in order to perform comparable to standard convolution on small-scale images. Therefore, in reality, $\hat{M}$ is much less than $M$ and so does $\hat{N}$. For instance, images can be represented by 32 channels in standard convolution rather than 144 or even larger in MobileNet-V2. In this situation, the reduction ratio is $\frac{M}{\hat{M} \cdot \hat{N}} + \frac{MN}{D^2_K \cdot \hat{M} \cdot \hat{N}} = \frac{144}{32 \cdot 32} + \frac{144 \cdot 144}{3^2 \cdot 32 \cdot 32}=2.39>1$. It indicates a standard convolution can even save more than half of computation cost compared with MobileNet-V2. Hence, it is reasonable to select the standard convolution layer for small size image and model.

\begin{figure*}[t]
	\centering
	\includegraphics[width=0.8\textwidth]{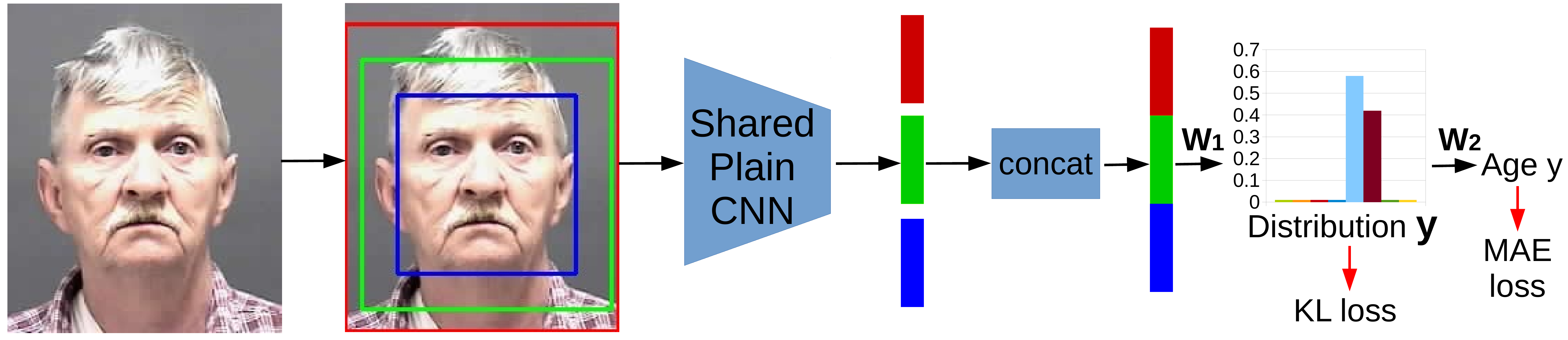}\\
	\caption{Overview of our compact model on age estimation.}
	\label{Fig: pipline}
\end{figure*}

\subsection{Two-Points Representation of Age}
In this section, we present a novel age representation as a distribution over two discrete adjacent bins.
Given a set of images $\{{(\matr{I}_n, y_n)}\}_{n = 1, 2, \cdots, N }$, deep regression model can be written as a mapping $\mathcal{F}: \mathcal{I} \rightarrow \mathcal{Y}$, where $\matr{I}_n$ and $y_n$ represent image and regression label, respectively. For any regression label $y_n$, it can be represented as a convex combination of two other numbers $z_n^1$ and $z_n^2$ ($z_n^1 \neq z_n^2$),
\vspace{-5pt}
\begin{equation}\label{eq:two_point_representation}
\resizebox{.3\hsize}{!}
{$
y_{n} = \lambda_1 z_n^1 + \lambda_2 z_n^2,
$}
\end{equation}
where $\lambda_1$ and $\lambda_2$ are the weights, $\lambda_1, \lambda_2 \in \mathbb{R}^{+}$, $\lambda_1 + \lambda_2 = 1$. 

\begin{figure}[!ht]
	\centering
	\includegraphics[width=0.85\linewidth]{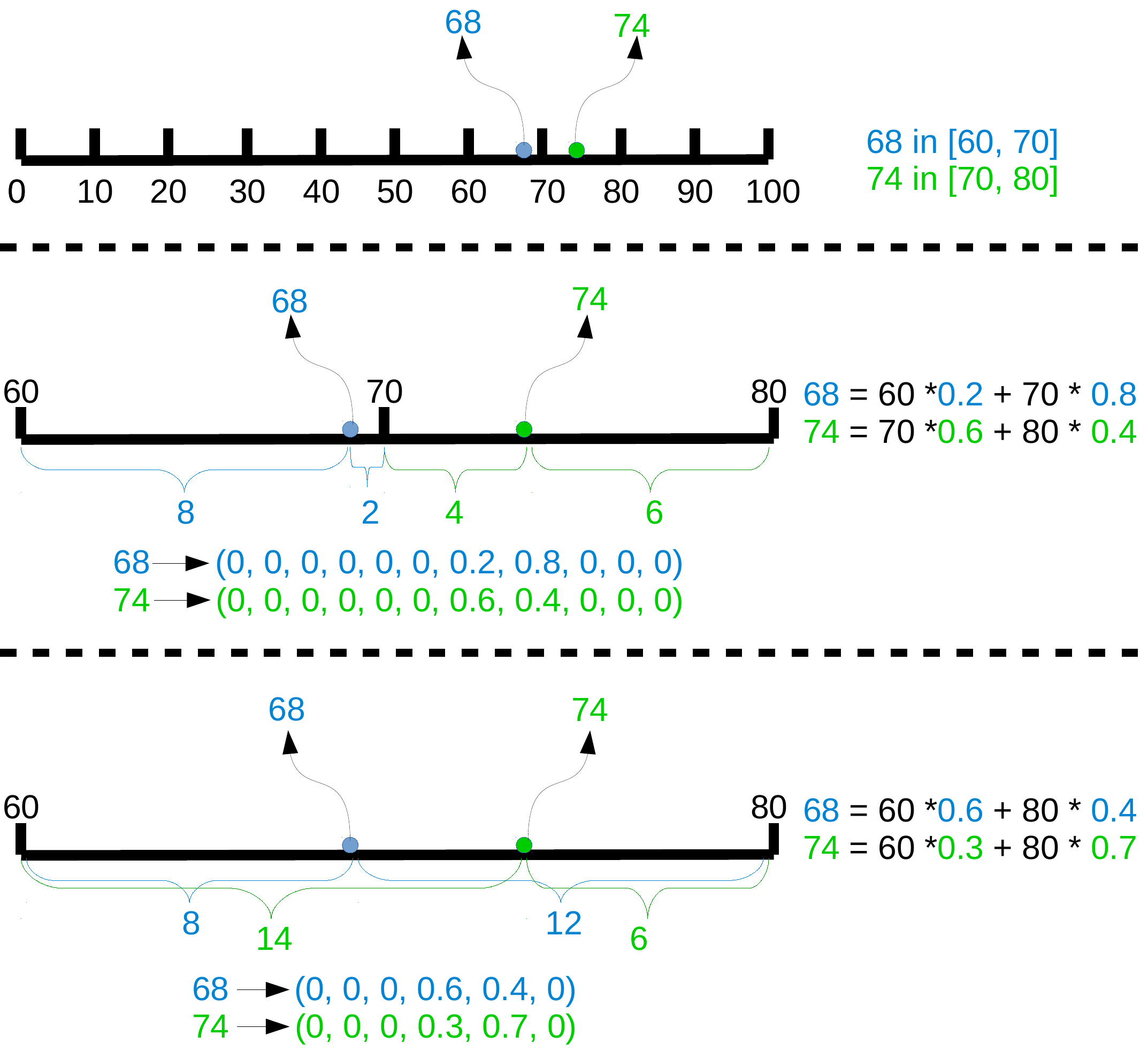}\\
	\caption{A new definition on the age estimation by two-points representation. Any point is represented by two adjacent bins instead of any other two or more bins.}
	\label{Fig: two_points_show}
\end{figure}

Given the age interval $[a,b]$, a label $y_n \in [a, b]$ and bins $\{z^m\}$ with uniform interval $K$ , $y_n$ can be represented by $z_n^1 = \left \lfloor \frac{y_n}{K} \right \rfloor \cdot K$ and $z_n^2 = \left \lceil \frac{y_n}{K} \right \rceil \cdot K$, where $\lfloor\cdot\rfloor$ and $\lceil\cdot\rceil$ are the floor and ceiling function. Accordingly, the coefficients $\lambda_1$ and $\lambda_2$ are computed as 
\vspace{-5pt}
\begin{equation}
\label{eq:lambda}
\resizebox{.6\hsize}{!}
{$
\begin{aligned}
\lambda_1 = & 1 - \frac{y_n - z_n^1}{K} = 1 - \frac{y_n - {\left \lfloor \frac{y_n}{K} \right \rfloor \cdot K}}{K} \\
\lambda_2 = & 1 - \frac{z_n^2 - y_n}{K} = 1 - \frac{{\left \lceil \frac{y_n}{K} \right \rceil \cdot K} - y_n}{K}
\end{aligned}
$}
\end{equation}
For example, as shown in Fig.~\ref{Fig: two_points_show}, the corresponding representation of 68 or 74 with $K = 10$ (second row in Fig.~\ref{Fig: two_points_show}) or $K=20$ (third row in Fig.~\ref{Fig: two_points_show}) is given. If $K = 10$, the set of bins is $\{10, 20, 30, 40, 50, 60, 70, 80\}$ and $y_n$ is 68, the corresponding vector representation is $\mathbf{y_n} = [0, 0, 0, 0, 0, 0.2, 0.8, 0]$. 
This operation assigns a distribution to the label, and will not incur any additional cost on distribution labeling. Moreover, the distribution of two-points representation is sparse. 

In fact, $\lambda_1$ and $\lambda_2$ represent the probability belonging to two bins, which include  rich distribution information. The main trend on age estimation includes two aspects: simultaneously classification and regression, and distribution learning. For the former, according to the above Fig.~ \ref{Fig: two_points_show}, 68 more likely belongs to bin 70 instead of bin 60. Two-points representation can disambiguate this problem naturally.
For the latter, some methods \cite{geng2013label, gao2017deep, pan2018mean} use distribution matching for better results. However, that requires extensive labeling to obtain the distribution that is very costly.

What is more, two-points representation gets two adjacent bins instead of any other two or more points, and the two adjacent bins are assigned with nonzero elements. In fact, each point/age in the linesegment can be represented by multiple points in which the number of combinations is very diversified. Each point can also be represented by two points or any other more points. However, those combinations is probably not what we want, e.g., $50=0.5 \times 0+0.5 \times 100=0.2 \times 10 + 0.2 \times40 + 0.2 \times 60 + 0.2 \times 90$. For age estimation, these representation is useless. While for deep regression model, these combinations need to be eliminated.

\subsection{Cascade Training}
From the above section, age value $y_n$ can be represented as distribution vector $\mathbf{y_n}$. However, the combination of $\mathbf{y_n}$ is diversified. Two-points representation is suitable to control it. The next question is how to embed the vector information into an end-to-end network. 
We implement this step by the cascade model shown in Fig. \ref{Fig: pipline}. In specific, a fully connected layer with semantic distribution is inserted in between feature layer $\mathbf{y_n}$ and the regression layer $y_n$. The mapping $f$ from feature $\mathbf{X}$ to age value $y$ can be decomposed into two steps $f_1$ and $f_2$, i.e., $f = f_2 \circ f_1$. In fact, the whole process can be denoted as $f: \matr{I}_n     \xrightarrow{Conv} \mathbf{X} \xrightarrow{\textbf{W}_1} \mathbf{y_n} \xrightarrow{\textbf{W}_2} y_n$.

Here we define two losses for two cascade task. The first one measures discrepancy between ground-truth label and predicted age distribution. We adopt KL-Divergence as the measurement,
\begin{equation}
\label{eq:KL_loss}
\resizebox{.7\hsize}{!}
{$
\begin{aligned}
L_{kl} (\textbf{y}_{n}, \hat{\textbf{y}}_{n}) & = \sum_n D_{KL}(\textbf{y}_{n}|\hat{\textbf{y}}_{n}) + \lambda ||\textbf{W}_1||_1 \\
& = \sum_n \sum_k \textbf{y}_{n}^{k} log \dfrac{\textbf{y}_{n}^{k}}{\hat{\textbf{y}}_{n}^{k}} + \lambda ||\textbf{W}_1||_1,
\end{aligned}
$}
\end{equation}
where $\textbf{W}_1$ is the weight of the mapping $f_1$ from concatenated feature $\mathbf{X}$ to the distribution $\hat{\textbf{y}}_{n}$, $\lambda$ is used to control the sparsity of the $\hat{\textbf{y}}_{n}$.
The second loss controls the prediction of the final age and is implemented as L1 distance (MAE loss),
\vspace{-5pt}
\begin{equation}
\label{eq:reg_loss}
\resizebox{.55\hsize}{!}
{$
\begin{aligned}
L_{reg} (y_n, \hat{y}_n) & = \sum_n ||y_n - \hat{y}_n||. \\
\end{aligned}
$}
\end{equation}

In the training process, two loss functions are trained in cascade style as shown in Fig.~\ref{Fig: pipline}. However they are still trained jointly, and the total loss is given as 
\vspace{-4pt}
\begin{equation}
\label{eq:total_loss}
\resizebox{.38\hsize}{!}
{$
\begin{aligned}
L_{total} & = \alpha L_{kl} + L_{reg} \\
\end{aligned}
$}
\end{equation}
where $\alpha$ is the hyperparameter to balance two losses. The cascade training can properly control the distribution $\hat{\mathbf{y}}_n$ in case of diversified combination.

\subsection{Context-based Regression Model}
The resolution and the size of small-scale image is limited. Exploiting facial information at different granularity levels is necessary. As shown in Fig.~\ref{Fig: motivation}, each cropped image has a special view on the face. The image with high resolution contains rich local information, in return one with low resolution may contain global and scene information.
Other than selecting one aligned facial center in SSR~\cite{yang2018ssr}, we crop face centers with three granularity levels, as shown in Fig.~\ref{Fig: pipline}, then fed them into the shared CNN network. Finally the bottlenecks of three-scale facial images are aggregated by concatenation that followed by cascade module.

\subsection{Discussions}
In this section, we summarize two non-trivial empirical guidelines for small-scale images and models. We will support our claims by experiments in the next section.

\textbf{Residual module} For small-scale image and model, is the residual module necessary? At least for age estimation dataset, it is not. Residual module with shortcut strategy is first designed by \cite{Kaiming2015Deep} to solve the problem of gradient vanishing, especially on very deep network. Its shortcut power can only be disclosed when enough layers were involved. The small-size model usually includes only shallow layers. According to our experiment, common connection on plain convolution is enough for small image and model. This discussion reminds us to rethink the apparent ideas in deep learning, especially on the small size image and model.

\textbf{SE module} The squeeze-and-excitation (SE) module has been validated by many works \cite{sandler2018mobilenetv2, ma2018shufflenet} for large scale image. While for small size image and model it also works well. So we integrate the SE module into our network and it costs very few parameters. For example, when the squeeze factor is 2, each SE module's parameters is only 32*16 *2 = 1024.

\section{Experiments}
The experiments consist of three parts. The first part is ablation study I on the comparison among SSR, MobileNet-V2, ShuffleNet-V2 and C3AE using plain model. The second one gives ablation study II on necessity of cascade module and context based module. The last one mainly provides the comparison with some state-of-the-arts.

\subsection{Datasets}
We study age estimation on three datasets: IMDB-WIKI~\cite{Rothe2016Deep}, Morph II~\cite{ricanek2006morph} and FG-NET~\cite{fu2010age}. 
We follow the conventions in the literature SSR~\cite{yang2018ssr}, DEX~\cite{Rothe2016Deep} and Hot~\cite{Rothe2016Deep}, WIKI-IMDB are used for pre-training and the ablation study. Because Morph II is the most popular and large benchmark for age estimation, we choose it for ablation studies.  Morph II and FG-NET are used to compare with the state-of-the-arts. 

\textbf{IMDB-WIKI} is the largest facial dataset with age labels, which is introduced in~\cite{Rothe2016Deep} and consists of $523,051$ images in total. The range is from $0$ to $100$. It is separated as two parts: IMDB($460,723$ images) and WIKI ($62,328$ images). However, it is not suitable for the performance evaluation on the age estimation because it contains much more noise. Thus, following previous works, e.g., SSR~\cite{yang2018ssr} and DEX~\cite{Rothe2016Deep}, we utilize IMDB-WIKI only for pre-training. 

\textbf{Morph II} is the most popular benchmark for age estimation, which has around $55,000$ face images of $13,000$ subjects with age label. The age ranges from $16$ to $77$(on average, $4$ images per  subject). Similar to some previous works~\cite{niu2016ordinal,zhang2017quantifying}, we randomly partition the dataset into two independent parts: training ($80\%$) and testing ($20\%$).

\textbf{FG-NET} contains $1,002$ face images from $82$ non-celebrity subjects with large variation of lighting, pose, and expression. The age ranges from $0$ to $69$ (on average, $12$ images per subject)~\cite{fu2010age}. Since the size of FG-NET is small, some previous methods usually use leave-one-out setting which needs to train 82 deep models. Under this setting, there are about 12 samples for the testing. Here we randomly choose 30 samples as the testing set and the remaining ones are for the training.
 We repeat this split $10$ times and compute their average performance.

\subsection{Implementation Details}
Following SSR \cite{yang2018ssr} and DEX \cite{Rothe2016Deep},
the model is firstly pre-trained on the IMDB and WIKI dataset, and is with size of $64 \times 64 \times 3$. In all the experiments, Adam optimizer is employed.
In the first ablation study, because the plain model of C3AE is compared with other plain models, each model is trained $160$ epochs with batch size of $50$. Similar to SSR, the initial learning rate, dropout rate, the momentum and the weight decay are set to $0.002$, $0.2$, $0.9$ and $0.0001$, respectively. The learning rate is decreased by a factor of the regression value with patience epochs $10$ on the change value of 0.0001. In the second ablation study, for comparing with the state-of-the-art methods, each model is trained $600$ epochs in total with the batch size of $50$. We use the strategy in \cite{zhong2017random} with randomly dropping out blocks on the input image. In this phase, the initial learning rate, dropout rate, the momentum and the weight decay are set to $0.005$, $0.3$, $0.9$ and $0.0001$, respectively. The learning rate is decreased by a factor of the regression value with patience epochs $20$ on the change value of 0.0005. Following SSR \cite{yang2018ssr}, the evaluation criteria is mean absolute value (MAE). The factor $\alpha$ in Eq.~\ref{eq:total_loss} is set to 10 in all the experiments. For all the cascade model, $K$ in Eq.~\ref{eq:lambda} is set to 10.

\subsection{Ablation Study}
The ablation study is conducted as two parts. For the first one, our plain model is compared with SSR, MobileNet-V2 and ShuffleNet-V2 to demonstrate that standard convolution yields competitive performance, even better than fashionable models such as MobileNet-V2 and ShuffleNet-V2. We further study whether the residual module and SE module can benefit small network.
For the second part, we conduct ablation study on the necessity of two-points representation and context module.

\vspace{-8pt}
\subsubsection{Ablation Study I: the Plain Model of C3AE}
This part includes three groups of experiments: comparison among our plain model, SSR, MobileNet-V2 and ShuffleNet-V2; comparing with/without residual module; and comparing with/without SE module.

The results of three methods (SSR, MobileNet-V2 and ShuffleNet-V2) on Morph II(M-MAE), IMDB (I-MAE) and WIKI (W-MAE) are given in Tab.~\ref{Tab: SSR_mobile_shuffle}.
For fair comparison, we implement extensive factor combinations(Comb.). In Tab.~\ref{Tab: SSR_mobile_shuffle}, for MobileNet-V2 (M-V2)\footnote{The code is from keras application}, $(\alpha_{pw}$, $\alpha_{exp})$ means the number of the pointwise filters and the expansion factor for each expansion layer, respectively. For ShuffleNet-V2 (S-V2)\footnote{The code is from https://github.com/opconty/keras-shufflenetV2}, $(\alpha_{ra}$, $\alpha_{fa})$ means ratio of bottleneck module's output channels for each stage and the scale factor for each stage's output channels, respectively. To conclude from the comparison, our plain model even with minimal parameters(Param.) and memory achieved best result regardless of parameter tuning in the alternative three methods.

We also give a speed analysis from two points: MACC and runtime speed. The former is the theoretical number of multi-add operations. The latter is the measured speed all under the same condition (forward single image 2000 times and then average), on CPU(Intel Xeon 2.1GHZ) and GPU(Titan X). The comparison is shown in Tab.~\ref{Tab: speed}.\\

\begin{figure*}[htb]
	\centering
	\includegraphics[width=1.02\textwidth]{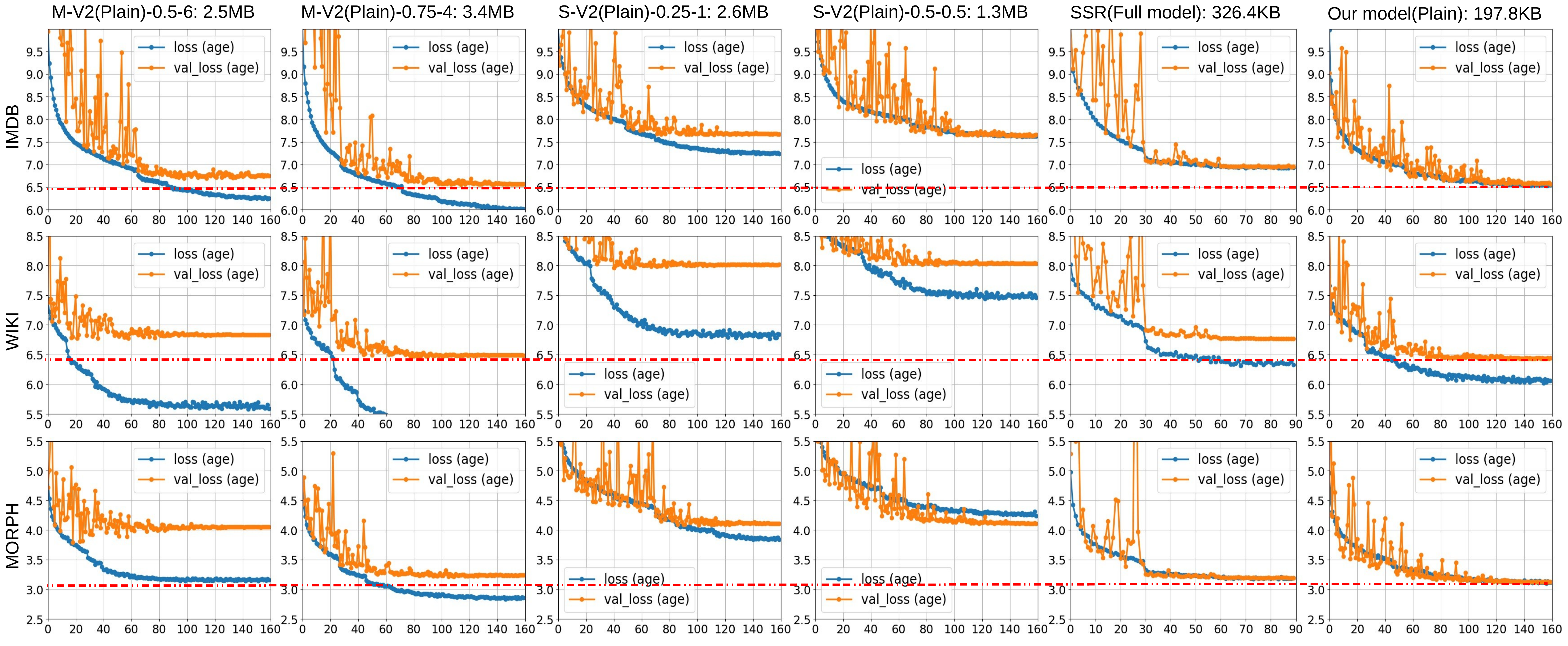}\\
	\caption{Comparison on the training process of M-V2, S-V2, SSR and our plain model.(Best viewed in color and magnifier.)}
	\label{Fig: training_testing_mobile2_shuffle2}
\end{figure*}

\vspace{-8pt}
As shown in Fig.~\ref{Fig: training_testing_mobile2_shuffle2}, the plain model of C3AE is consistently better than SSR, ShuffleNet-V2 and MobileNet-V2 with lower validation loss (val\_loss in orange, training loss in blue). More examples can be found in the supplementary material. For MobileNet-V2 and ShuffleNet-V2, with the depth-wise convolution, is by no means inferior than our plain model with standard convolution. 
In addition, there is a strange observation that the result of $\alpha_{exp} = 4$ is superior to  $\alpha_{exp} = 6$. We believe that too large inverted bottleneck may be not suitable for small size model. For SSR, the standard convolution is also used. However, its full model is still inferior to our plain model. In addition, the gap between train and validate loss in our plain is the least. It shows our plain model has better generalization.
All these observations suggest the effectiveness of our plain model. Although our plain model is plain enough without any bells and whistles, it still can get very competitive performance.

We further investigate the effectiveness of residual connection and SE module. According to the results in Tab.~\ref{Tab: wo_residual_module_and_SE} and the comparison in supplementary material, we observe that residual module does not benefit in the small size model, in particular for three datasets on age estimation. While SE module work well for small size model.

\vspace{-2pt}
\begin{table}[!htbp]
	\setlength{\abovecaptionskip}{5pt}
	\footnotesize
	\begin{center}
		\caption{Comparsion among SSR, M-V2, S-V2 and C3AE.} %
		\tabcolsep=1pt
		\begin{tabular}{c|c|c|c|c|c|c|c}
			\hline
			\hline
			Methods & Comb.   & M-MAE & I-MAE & W-MAE & Param. & Memory & MACC\\
			\hline
			\multirow{6}[2]{*}{M-V2} & (0.25, 4) & 3.72 & 7.23 & 7.29 & 107129 & 808.7KB & 2.2M \\
			& (0.25, 6)  & 4.26 & 7.01 & 7.30 & 153561 & 994.7KB & 3.0M \\
			& (0.5, 4)    & 3.71 & 6.76 & 6.76 & 354713 & 1.8MB & 5.7M \\
			& (0.5, 6)   & 4.05 & 6.75 & 6.83 & 518857 & 2.5MB & 8.1M \\
			& (0.75, 4)  & 3.24 & \textbf{6.57} & 6.49 & 747961 & 3.4MB & 12.3M\\
			& (0.75, 6)   & 4.10 & 6.69 & 6.72 & 1102537 & 4.8MB & 17.7M \\
			\hline
			\multirow{6}[2]{*}{S-V2} & (0.25, 0.5)                               & 4.85 & 8.22 & 8.78 & 76589 & 1.0MB & \textbf{0.6M}\\
			& (0.25, 1)   & 4.11 & 7.67 & 8.02 & 464185& 2.6MB & 4.0M\\
			& (0.5, 0.5)  & 4.11 & 7.66 & 8.04 & 155753& 1.3MB & 1.4M\\
			& (0.5, 1)    & 3.83 & 7.40 & 7.63 & 1284087& 5.9MB & 12.7M\\
			& (0.75, 0.5) & 3.98 & 7.55 & 7.91 & 250829& 1.7MB & 2.5M\\
			& (0.75, 1)   & 3.63 & 7.07 & 7.19 & 2473043& 10.7MB & 26.1M\\
			\hline
			SSR & Full model & 3.16 & 6.94 & 6.76 & 40915 & 326.4KB & 17.6M \\
			\hline
			C3AE & Plain model & \textbf{3.13} & \textbf{6.57} & \textbf{6.44} & \textbf{36345} & \textbf{197.8KB}\ & 12.8M\\
			\hline
		\end{tabular}%
		\label{Tab: SSR_mobile_shuffle}
	\end{center}
\end{table}

\vspace{-5pt}
\begin{table}[!htbp]
	\footnotesize
	\begin{center}
	\vspace{-8pt}
		\caption{The Speed analysis} %
		\tabcolsep=0.5pt
		\begin{tabular}{c|c c c c c c}
			\hline				
			evaluation	   & our-plain   & SSR   & M-v2(.5,6) & M-v2(.75,6) & S-v2(.5,1) & S-v2(.75,1)\\
			\hline
		     MACC (M)      & 12.8  & 17.6  & \textbf{8.1} & 17.7      &12.7 &26.1                  \\
		     runtime-cpu(s)& \textbf{0.0126}& 0.0233& 0.0245      &0.0394         &0.0228 &0.0295 \\
		     runtime-gpu(s)& \textbf{0.0029}& 0.0050& 0.0070      &0.0080         &0.0080 &0.0082          \\
		     \hline
		     MAE           & 3.13  & 3.16  & 4.05      & 4.10 & 3.83 & 3.63 \\
			\hline
		\end{tabular}
		\label{Tab: speed}
	\end{center}
\end{table}

\begin{table}[htbp]
	\setlength{\belowcaptionskip}{-8pt}
	\footnotesize
	\begin{center}
	\vspace{-5pt}
		\caption{The role of residual module and SE} %
		\begin{tabular}{c|c c c}
			\hline
			\hline				
			Datasets	  & w/o Res+w/o SE & w. Res  & w. SE \\
			\hline
		     Morph II     & 3.13           & 3.21   & 3.11\\
		     IMDB         & 6.57           & 6.66   & 6.50\\
		     WIKI         & 6.44           & 6.57   & 6.36\\
			\hline
		\end{tabular}
		\label{Tab: wo_residual_module_and_SE}
	\end{center}
\end{table}

\vspace{-12pt}
\subsubsection{Ablation Study II: Cascade and Context Module}
In this section, we analyze how the choice of cascade module (two-points representation)  and context module affect the performance of age estimation.  

\begin{figure}[!htb]
    \centering
    \includegraphics[width=0.45\textwidth]{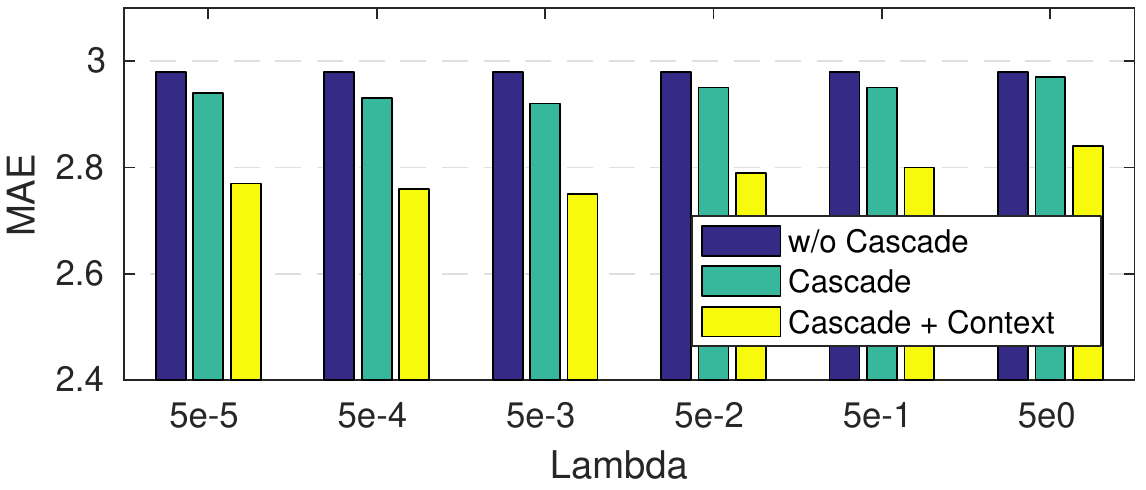}
   \vspace{-5pt}
    \caption{Evaluation of cascade and context module.}
    \label{fig:CascadeContext}
\end{figure}

The result of two-points representation is implemented by cascaded training, i.e., with/without cascade module. 
As shown in Fig.~\ref{fig:CascadeContext}, regardless of the regularizer $\lambda$ in Eq.~\ref{eq:KL_loss} we choose, the result with casacde module is consistently better than that without cascade. If the context module is further applied (Cascade + Context) it outperforms the other two.
The validations demonstrate the importance of two-points representation and context module. 

\begin{figure*}[htbp]
	\centering
	\includegraphics[width=1.02\textwidth]{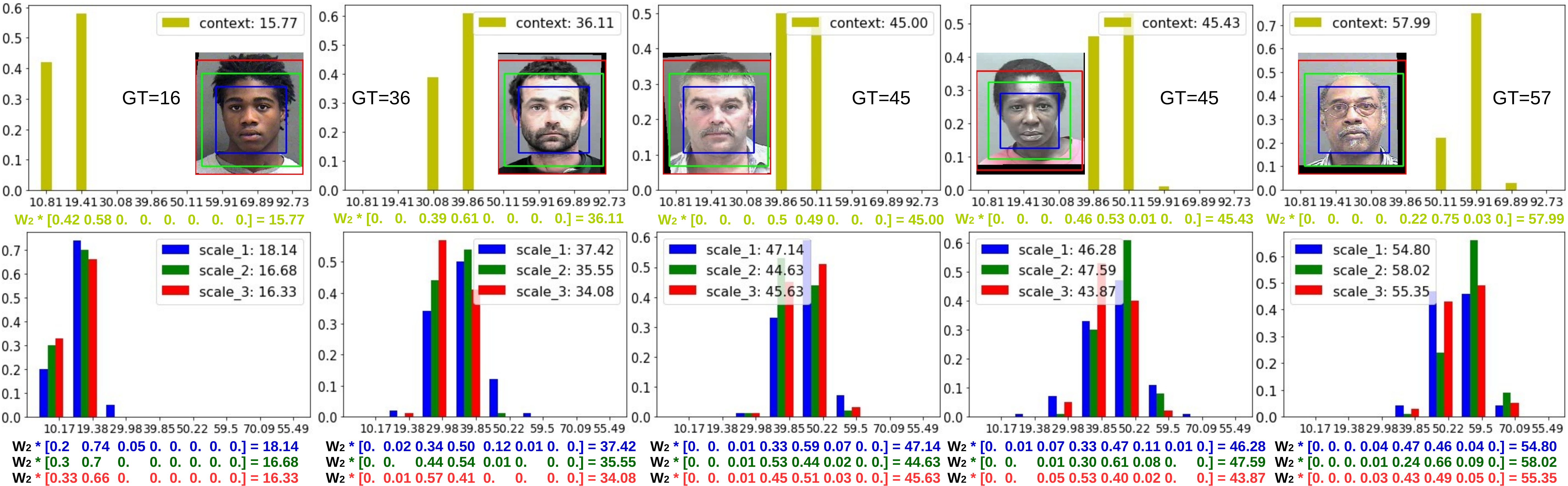}\\
	\caption{Some examples on the C3AE. Top: the result of the context based regression model. The yellow bars denote the predicted distribution $\hat{\textbf{y}}_{n}$, and the X-axis is the learned weight $\textbf{W}_2$ from the distribution to age value.  Bottom: Three different colors RGB correspond to each facial context and predicted distribution $\hat{\textbf{y}}_{n}$.(Best viewed in color and magnifier.)}
	\label{Fig: examples}
\end{figure*}

In specific, we give some examples in Fig.~\ref{Fig: examples}. GT means the groundtruth value, and the legend gives the predicted age. The X-axis is the learned weights $\textbf{W}_2$, and the Y-axis is the predicted vector $\hat{\textbf{y}}_{n}$. Their dot/inner product is the predicted age. We can see that the learned weights are almost equivalent to groundtruth bins $\textbf{W}_2=[10, 20, 30, 40, 50, 60, 70, 80]$. That is to say, $\textbf{W}_2$ controls two-points representation so that the diversified combinations are eliminated. The last element of the predicted bins is very strange, i.e., $92.73, 55.49$. After the analysis of the data distribution, we found that there are only 9 samples in the range [70, 80], and it is easy to explain why the last element is abnormal. The predicted distribution is sparse with only two or three adjacent nonzero elements because of two-points representation. 
Fully connected layer will lead to the phenomenon that each age can be represented by many different combinations. 

In addition, as shown in Fig.~\ref{Fig: examples}, we also observe that the predicted distribution and age on the top is better than that on the bottom. The colors of the bar, legend and the distribution correspond to the colored bounding box on the top image. Context based model (top) achieves better performance than that of single scale input (bottom). 

Finally, in order to show generality of our model, we finetune the hyperparameters $\alpha$ as 5, 8, 10, 12 and 15 on our full model, and the corresponding results are 2.79, 2.79, 2.75, 2.79 and 2.80, respectively. These results does not change too much. It shows the robustness of our model.

\subsection{Comparison with State-of-the-arts on Morph-II}
In this section, we further compare our model with state-of-the-art models on Morph II. As shown in Tab.~\ref{Tab: state_of_art_morph}, our full model achieves 2.78 and 2.75 MAE under the condition: trained from scratch and pretrained on IMDB-WIKI, which is the state-of-the-art performance among compact models. The previous best performance achieved in the compact model is $3.16$ in SSR~\cite{yang2018ssr}. Some results in the Tab.~\ref{Tab: state_of_art_morph} are from SSR~\cite{yang2018ssr}. In fact, our plain model achieves 3.13 MAE even without any bells and whistles. The results of all other compact models are pretrained on IMDB-WIKI. Our results on with/without pretrained process are very similar. We believe Morph II is large enough to train our tiny model. On the other hand, our result is much competitive compared with the bulky models, and it even surpasses several bulky models despite it consumes only 1/2000 of their model sizes. All the bulky models are pretrained on ImageNet or IMDB-WIKI using VggNet. Our result without pretrained process even surpasses some pretrained bulky models. In general, C3AE gets very competitive performance on Morph II with extremely lightweight model.

\vspace{-2pt}
\begin{table}[htbp]
	\setlength{\abovecaptionskip}{5pt}
	\setlength{\belowcaptionskip}{-5pt}
	\footnotesize
	\begin{center}
		\caption{Comparsion with state-of-the-arts that use compact and bulky basic models on Morph II.} %
		\tabcolsep=1pt
		\begin{tabular}{c | c|c c c c}
			\hline	
			\hline			
			Type & Methods	  & MAE & Memory & Parameters\\
			\hline
			\multirow{5}[2]{*}{Compact} & ORCNN	\cite{niu2016ordinal}  & 3.27 & 1.7MB & 479.7K\\
			& MRCNN	\cite{niu2016ordinal}  & 3.42 & 1.7MB & 479.7K\\
			& DenseNet \cite{huang2017densely} & 5.05 & 1.1MB & 242.0K \\
			& MobileNet-V1 \cite{howard2017mobilenets} & 6.50 & 1.0MB & 226.3K\\
			& SSR	\cite{yang2018ssr}      & 3.16 & 0.32MB & 40.9K \\
\hline
\multirow{5}[2]{*}{Bulky} & Ranking CNN \cite{chen2017using}	& 2.96 & 2.2GB & 500M\\
			& Hot \cite{rothe2016some}   & 3.45 & 530MB & 138M\\
			& ODFL \cite{liu2017ordinal} & 3.12 & 530MB  &  138M\\
			& DEX \cite{Rothe2016Deep}  & 3.25 & 530MB & 138M\\
			& DEX (IMDB-WIKI) \cite{Rothe2016Deep}  & 2.68 & 530MB & 138M\\
			& ARN \cite{agustsson2017anchored}   & 3.00 & 530MB & 138M\\
			& AP \cite{zhang2017quantifying}	 & $2.52$ & 530MB & 138M\\%
			& MV \cite{pan2018mean}	 & $2.41$ & 530MB & 138M \\
			& MV (IMDB-WIKI) \cite{pan2018mean}	 & $\textbf{2.16}$ & 530MB & 138M\\%
\hline
\multirow{2}[2]{*}{C3AE} 
	& Full model (Scratch) & 2.78 & \textbf{0.25MB} & \textbf{39.7K} &  \\
	& Full model (IMDB-WIKI) & \textbf{2.75} & \textbf{0.25MB} & \textbf{39.7K} &  \\
			\hline
		\end{tabular}
		\label{Tab: state_of_art_morph}
	\end{center}
\end{table}

\vspace{-14pt}
\subsection{Comparison with State-of-the-arts on FG-NET}
As shown in Tab.~\ref{Tab: state_of_art_fg}, we compare our model with state-of-the-art models on FG-Net.
Without training 82 models, we randomly repeat the experiment ten times. This is also challenging because we use less train dataset. In fact, Han \cite{han2015demographic}, Luu \cite{luu2009age, luu2011contourlet} in Tab.~\ref{Tab: state_of_art_fg} are also use the different splits. 
 Using mean-variance loss, MV~\cite{pan2018mean} with pre-trained process gets the best result of 2.68. While our result with pre-trained process is 2.95 MAE and 0.17 std, i.e., the second best performance compared with Bulky models. In addition, without any pre-trained process, our result of 4.09 is slightly better than MV~\cite{pan2018mean} of 4.10. 
In general, the validation on FG-NET demonstrate the effectiveness of C3AE.

\begin{table}[htbp]
	\setlength{\abovecaptionskip}{3pt}
	\setlength{\belowcaptionskip}{-3pt}
	\footnotesize
	\begin{center}
		\caption{Comparison with state-of-the-arts on FG-Net.} %
		\tabcolsep=1pt
		\begin{tabular}{c|c c c c}
			\hline	
			\hline			
			Methods	  & MAE & Memory & Parameters\\
			\hline
		Geng~\etal~\cite{geng2013label}	  & 5.77 & - & - &  \\
		Han~\etal~\cite{han2015demographic}	  & 4.80 & - & - &  \\
		Luu~\etal~\cite{luu2009age}	  & 4.37 & - & - &  \\
		Luu~\etal~\cite{luu2011contourlet}	  & 4.12 & - & - &  \\
		Wang~\etal~\cite{wang2015deeply} & 4.26 & - & -\\
		Feng~\etal~(1)~\cite{feng2017human} & 4.35 & 530MB & 138M\\
		Feng~\etal~(2)~\cite{feng2017human} & 4.09 & 530MB  & 138M\\
		Zhu~\etal (Actual)~\cite{zhu2018facial} & 4.58 & 530MB  & 138M\\
		Zhu~\etal (Synthesized)~\cite{zhu2018facial} & 3.62 & 530MB  &  138M\\
		Liu~\etal~\cite{liu2017ordinal} & 3.89 & 530MB  &  138M\\
		DEX~\cite{Rothe2016Deep}     & 4.63 &  530MB &  138M\\
		DEX (WIKI-IMDB)     \cite{Rothe2016Deep}      & 3.09 &  530MB &  138M\\
		MV \cite{pan2018mean}	 & $4.10$ & 530MB & 138M \\
		MV (WIKI-IMDB) \cite{pan2018mean}	 & \textbf{$\textbf{2.68}$} & 530MB & 138M\\%
		\hline
		C3AE (Scratch)	  & $4.09\pm 0.19$ & \textbf{0.25MB} &  \textbf{39.7K}\\
	    C3AE	(WIKI-IMDB)  & $2.95\pm 0.17$ & \textbf{0.25MB} &  \textbf{39.7K}\\
	     \hline
		\end{tabular}
		\label{Tab: state_of_art_fg}
	\end{center}
\end{table}

\vspace{-15pt}
\section{Conclusion}
\vspace{-2pt}
In this paper, we have proposed a compact model, C3AE, that has achieved state-of-the-art performance among compact models and competitive performance among bulky models. From various ablation study, we have demonstrated the effectiveness of C3AE. For the small/medium-size image and model, some analysis and rethinking are given. In the future work, we will evaluate the effectiveness of our observation on general datasets and applications.

\vspace{-6pt}
\section{Acknowledgements}
\vspace{-4pt}
This research was supported in part  by National Natural Science Foundation of China (NSFC, No.61571102, No.61602091, No.61872067),  Research Programs of Science and Technology in Sichuan (No.2018JY0035, No.2019YFH0016).

{\small
\bibliographystyle{ieee_fullname}
\bibliography{reference}
}

\end{document}